\pdfoutput=1
\documentclass[11pt]{article}

\usepackage[]{acl}

\usepackage{times}
\usepackage{latexsym}
\usepackage[]{mdframed}
\usepackage{booktabs}
\usepackage{multirow} 
\usepackage[T1]{fontenc}
\usepackage{authblk}

\usepackage[utf8]{inputenc}

\usepackage{microtype}

\usepackage{inconsolata}
\usepackage{graphicx}
\usepackage{lscape}

\title{Towards Large Language Model driven Reference-less Translation Evaluation for English and Indian Languages}

\author[ ]{Vandan Mujadia}
\author[ ]{Pruthwik Mishra}
\author[ ]{Arafat Ahsan}
\author[ ]{Dipti Misra Sharma}
\affil[ ]{LTRC, IIIT Hyderabad, India}
\affil[ ]{\{vandan.mu, pruthwik.mishra\}@research.iiit.ac.in,}
\affil[ ]{ \{arafat.ahsan, dipti\}@iiit.ac.in}

\begin{document}
\maketitle
\begin{abstract}
With the primary focus on evaluating the effectiveness of large language models for automatic reference-less translation assessment, this work presents our experiments on mimicking human direct assessment to evaluate the quality of translations in English and Indian languages. We constructed a translation evaluation task where we performed zero-shot learning, in-context example-driven learning, and fine-tuning of large language models to provide a score out of 100, where 100 represents a perfect translation and 1 represents a poor translation. We compared the performance of our trained systems with existing methods such as COMET, BERT-Scorer, and LABSE, and found that the LLM-based evaluator (LLaMA-2-13B) achieves a comparable or higher overall correlation with human judgments for the considered Indian language pairs (Refer figure \ref{fig:mainfig}). 

\end{abstract}

\section{Introduction}

The field of natural language processing (NLP) and artificial intelligence (AI) has been transformed by the rapid advancements in Large Language Models (LLMs), as they have demonstrated their capabilities in a wide range of natural language processing tasks, including open/close question answering, summarization \citep{chang2023survey, min2023recent}, code completion, and code debugging \citep{wang2023survey, zan-etal-2023-large, surameery2023use}, etc. These models have significantly impacted NLP applications by enhancing various aspects of language understanding, generation, and analysis \citep{zhao2023survey}. As LLMs continue to evolve and improve, they hold immense potential for further advancements in NLP, paving the way for more sophisticated and intelligent solutions \citep{hadi2023large}. \\

Translation, on the other hand, plays a vital role in bridging the gap between different languages, enabling effective communication. With the advancement of machine translation systems \citep{ranathunga2023neural, gala2023indictrans2}, there has been a significant shift in how translations are produced \citep{arivazhagan2019massively}. Translation engines have made great progress in understanding languages and generating translations, but they still face numerous challenges in accurately conveying the intended meaning \citep{al2022taxonomy, 10.1162/coli_a_00435, hayakawa-arase-2020-fine}. While these systems have made remarkable progress, their evaluation remains an indispensable component in ensuring the accuracy and quality of translations produced by them \citep{sangal2022evaluating}.\\

\begin{figure}
    \centering
\includegraphics[width=.5\textwidth]{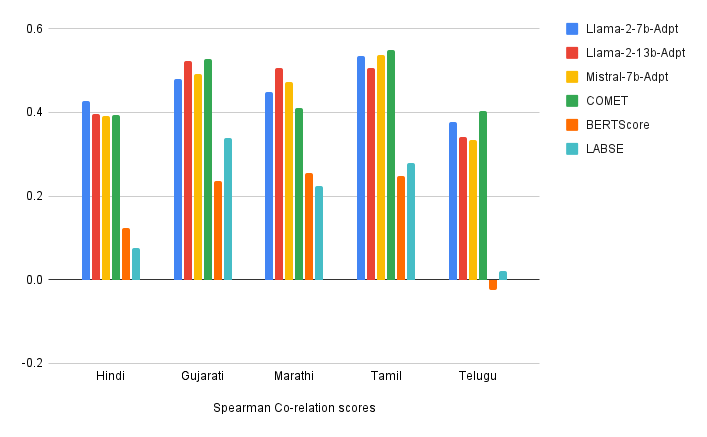}
    \caption{Spearman co-relation: Human translation evaluation vs different reference-less translation evaluation metrics. Llama-2-7b-Adapt (lora), Llama-2-13b-Adapt (lora), Mistral-7b-Adpt (lora), COMET-QE (\url{https://github.com/Unbabel/COMET}) are fine-tuned models using WMT-23 MT-QE corpora and evaluated on development corpora. BERTScore (\url{https://github.com/Tiiiger/bert_score}) and LABSE (\url{https://huggingface.co/sentence-transformers/LaBSE}) represents direct cosine-similarity scores.}
    \label{fig:mainfig}
\end{figure}
To date, human involvement has been considered one of the most reliable and effective ways to evaluate translations \citep{freitag2021experts,rivera2022machine,guzman-etal-2015-humans}. However, this requires a significant investment of time, expertise, and financial resources. The cost and effort involved can limit the frequency and scale at which human involved evaluation can be conducted. Therefore, it is crucial to develop a scalable, replicable, and efficient automatic method that mimics human evaluation to ensure reliable and effective translation assessment.\\

Recent developments in the field of automatic translation evaluation have demonstrated that techniques utilizing multilingual embeddings have a tendency to outperform other traditional approaches and display the strongest correlation with human assessments \citep{zerva-etal-2022-findings}. Notable instances of these techniques include BERTScore \citep{zhang2019bertscore} and COMET \citep{rei-etal-2020-comet,rei-etal-2022-cometkiwi,kocmi-etal-2022-ms}. To further improve the effectiveness of automatic translation evaluation, it is reasonable to investigate methodologies that leverage large language models, considering their notable capability for comprehension.\\

In this work, we aim to assess the capability of LLMs and utilize them for \textbf{reference-less translation evaluation} involving English and Indian languages. The research questions that we pose are as follows:
\begin{itemize}
    \item Do LLMs possess zero-shot or in-context translation evaluation capabilities?
    \item How do fine-tuned LLMs compare with existing state-of-the-art translation evaluation methods such as COMET, BertScore, and LABSE?
    \begin{itemize}
        \item Does translation evaluation fine-tuning for LLMs enhance their translation evaluation capabilities? Does fine-tuning improve their performance when applied to other unseen Indian languages in fine-tuning?
        \item Does fine-tuning LLMs for both translation and translation evaluation increase their evaluation capabilities?
    \end{itemize}

\end{itemize}

In this study, our objective is to gain a clearer understanding of the reference-less translation evaluation capabilities of several popular large language models. Specifically, we focus on Opt \citep{zhang2022opt}, bloom \cite{scao2022bloom}, LLaMA-1 \footnote{\url{https://huggingface.co/decapoda-research/llama-7b-hf}}, MPT \footnote{\url{https://huggingface.co/mosaicml/mpt-7b}}, Falcon \citep{refinedweb}, LLaMA-2 \cite{touvron2023LLaMA2}, and Mistral \citep{jiang2023mistral} for translating English to five Indian languages: Hindi, Gujarati, Marathi, Tamil, and Telugu. \\

Reference-less translation evaluation generally involves assessing the translation of a given source language sentence without a specific reference translation. In this evaluation, an automatic reference-less evaluation system assigns a score to the translation, which is similar to the direct assessment (DA) score \citep{akhbardeh-etal-2021-findings}, commonly used in human translation evaluation. This score ranges from 1 to 100, with 1 representing a poor translation and 100 representing a perfect translation. In our work, we initially evaluate the reference-less translation evaluation capabilities of the raw large language models mentioned above in both zero-shot \citep{NEURIPS2022_8bb0d291} and in-context learning \footnote{\url{https://ai.stanford.edu/blog/understanding-incontext/}} \citep{brown2020language} scenarios. Subsequently, we employ the LoRa parameter-efficient fine-tuning technique \citep{hu2021lora}, as well as full fine-tuning, to refine the selected base LLM models. Additionally, considering that LLMs are known for their multi-task learning abilities \citep{radford2019language}, we explore the fine-tuning of LLMs on translation corpora alongside translation evaluation corpora to determine if translation-based fine-tuning improves overall translation evaluation performance.\\

The key findings of our study, presented in Figure \ref{fig:mainfig}, highlight the performance of our fine-tuned LLM-based reference-less translation evaluation models compared to various well-known reference-less evaluation methods, such as COMET, LABSE, and Bert-scorer, based on their Spearman correlation with human judgments.\\

Our findings emphasize the significant potential of large language models for reference-less translation evaluation tasks involving English and Indian languages. Raw LLMs do not inherently possess the capabilities for translation evaluation as they do not provide a score as an evaluation outcome. However, our multi-lingual LLM based LORa-fine-tuned models (LLaMA-2-7b, LLaMA-2-13b and Mistral-7b) demonstrate competitive or superior correlation with human judgments compared to existing reference-less methods like COMET under same training and evaluation configurations. We did not observe any additional benefits when we perform translation evaluation fine-tuning with translation fine-tuning under the multi-task setting.\\

The results suggest that fine-tuned LLMs hold promise for translation evaluation in the targeted reference-less translation evaluation task. Our study represents an essential and pioneering milestone in assessing and enhancing the reference-less translation evaluation capabilities of LLMs, involving English and Indian languages.

\section{Related Work}
There are two main categories of automatic machine translation (MT) metrics: string-based metrics and pretrained models based metrics. In the following sub-section, we discuss them briefly one by one.

\subsection{String-based metrics:}
String-based metrics involve comparing the coverage of various substrings between the human-generated reference and MT translations. This includes metrics such as ChrF \citep{popovic2015chrf}, BLEU \citep{papineni2002bleu}, METEOR \citep{gupta2010meteor}, or TER \citep{snover2009ter}. String-based methods heavily rely on the quality of reference translations. However, they have the advantage of predictable and faster performance, as it is computationally easy to identify which substrings have the most impact on the score.

\subsection{Pretrained models based metrics:}
This category consists of metrics that utilize pretrained models to evaluate the quality of machine translation (MT) outputs given the source sentence, the human reference, or both. Evaluation metrics in this category include COMET \citep{rei-etal-2020-comet}, BLEURT \citep{sellam-etal-2020-bleurt}, and BERTScore \citep{zhang2019bertscore}. These models are based on pretrained models such as XLM-RoBERTa \citep{conneau-etal-2020-unsupervised} and MBERT \citep{DBLP:journals/corr/abs-1810-04805}. These metrics are not strictly dependent on reference quality and can better evaluate synonyms or paraphrases. Several studies \citep{mathur-etal-2020-tangled,kocmi-etal-2022-ms,akhbardeh-etal-2021-findings} have demonstrated their superiority over string-based metrics. However, it is important to note that the performance of these metrics is influenced by the data they have been trained on, which can introduce bias. Pretrained model-based metrics can be further categorized as reference-based and reference-free metrics. As the name suggests, reference-free metrics (also known as quality estimation metrics) do not require a human reference for evaluation. COMET-QE \citep{chimoto-bassett-2022-comet,rei-etal-2020-comet}, BERTScore \citep{zhang2019bertscore}, and LABSE \citep{feng-etal-2022-language} fall under this category of metrics. In this direction, we develop a reference-less metric utilizing the large language model (LLM) and human-labeled data for fine-tuning. In the following section, we discuss our considered LLM models briefly.%We use the same hyperparameters as COMET for consistency and comparability.

\section{Large Language Models}
\label{LLMb}
Language modeling, a well-established task in the field of natural language processing, has garnered significant attention over the years \citep{bellegarda2004statistical, bengio2000neural}. This task involves predicting the probability of the next token in a sequence of words. Transformers have emerged as the fundamental architecture underlying many existing Large Language Models \citep{vaswani2017attention}. Transformers based auto-regressive models like GPT \cite{brown2020language, radford2018improving, radford2019language} have played a crucial role in advancing Natural Language Processing (NLP). In this work, we used following base LLM models to check how we can utilise them for machine translation evaluation involving English and Indian Languages.\\
\textbf{opt-6.7b\footnote{\url{https://huggingface.co/facebook/opt-6.7b}}} : The OPT-6.7b \citep{zhang2022opt} model has been extensively trained on the objective of causal language modeling (CLM) using English text with 6.7 billion parameters.\\
\textbf{Bloom-7B\footnote{\url{https://huggingface.co/bigscience/bloom-7b1}}} : BLOOM \cite{scao2022bloom} was the first largest multilingual large language model with causal language modeling objective and supports 46 languages and 13 programming languages. It has 7,069,016,064 parameters.\\
\textbf{LLaMA-7B\footnote{\url{https://huggingface.co/decapoda-research/llama-7b-hf}}}: LLaMA is a collection of foundation language models ranging from 7B to 65B parameters. In our experiments we evaluated LLaMA model with 7B parameters where 4096 is the embedding dimensions and 32 layers and 32 attention head.\\
\textbf{MPT-7B\footnote{\url{https://huggingface.co/mosaicml/mpt-7b}}} : Similar to above models MPT-7B model is trained on a large amount of data 1T tokens on causal language modeling objective.\\
\textbf{Falcon\footnote{\url{https://huggingface.co/tiiuae/falcon-7b}}} : Falcon \citep{refinedweb} is another large language model trained on causal language modeling (CLM) objective. Here, we utilised Falcon-7B model which is a 7B parameters for our experiments. \\
\textbf{LLaMA-2-7B\footnote{\url{https://huggingface.co/meta-llama/Llama-2-7b-hf}} and LLaMA-2-13B\footnote{\url{https://huggingface.co/meta-llama/Llama-2-13b-hf}}} : LLaMA 2 based models \cite{touvron2023LLaMA2} are also trained on causal language modeling (CLM) objective and pretrained on 2 trillion tokens of data from publicly available sources. In our experiments we have experimented with 7B and 13B LLaMA-2 models. LLaMA-2-7B network has 32 layers and 32 attention heads while LLaMA-2-13B has 40 layers and 40 attention heads.\\
\textbf{Mistral-7B\footnote{\url{https://huggingface.co/mistralai/Mistral-7B-v0.1}}} : Mistral-7B Large Language Model (LLM) \citep{jiang2023mistral} is a pre-trained on causal language modeling (CLM) objective with 7 billion parameters. It uses Sliding Window Attention (SWA) to handle longer sequences at smaller cost and Grouped-query attention (GQA) for faster inference which reduces the memory requirement during decoding. It has 4096 embedding dimension, 32 layers and 32 attention heads with context length of 8192. 

\begin{figure*}[h]
    \centering
    \includegraphics[width=1.0\textwidth]{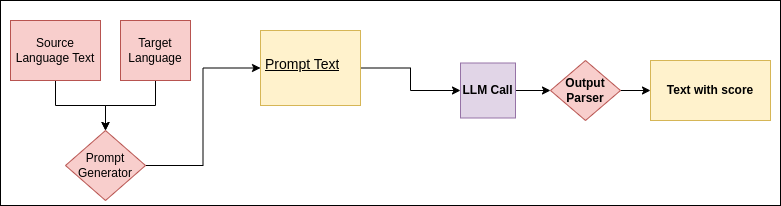}
    \caption{Prompting Mechanism for reference-less translation evaluation}
    \label{fig:prompting-1}
\end{figure*}

\section{Reference-less Translation Evaluation on Raw LLM}
To evaluate the effectiveness of the LLMs mentioned above for translation evaluation tasks without reference, we conducted two different experiments. The first involved assessing the performance of the pre-trained (raw) LLM for reference-less translation evaluation. In the second experiment, we performed example-based in-context learning for the same purpose. Both experiments were carried out using translation evaluation data mentioned in the section \ref{traindevingdata}.\\

As part of our experimental setup, we configured a prompting pipeline depicted in Figure \ref{fig:prompting-1}. This pipeline involved using a Prompt Generator to generate specific prompts for the source and target language pairs for source and translation text. Subsequently, an LLM call is triggered to generate a response, which was then processed by a reply parser to obtain the actual translation. To ensure high-throughput and memory-efficient inference and serving for LLMs, we utilized the vLLM library\footnote{\url{https://github.com/vllm-project/vllm}} \citep{kwon2023efficient}. We conducted all experiments using a temperature parameter of 0, which ensures that the model behaves deterministically. By setting the temperature to 0, the model is constrained to select the word with the highest probability, effectively limiting its choice to the most likely option \citep{aksitov2023characterizing}. All of our experiments are conducted using this library, and the models are deployed on A100, 40GB GPUs.

\subsection{Zero-shot}
We performed manual trials to determine the optimal prompt for zero-shot reference-less translation evaluation. These trials revealed that instructing an LLM to mimic a human evaluator and provide scores out of 100, combined with presenting the text in JSON format, resulted in better results (prompt presented below).

\begin{mdframed}
\textbf{Zero-shot Translation Evaluation Prompt Example:}\\
You are an experienced translation evaluator and you need to evaluate a translation for <Source Language> language to <Target Language> language. \\ 

<Source Language>: <Source Language Text> \\ 

<Target Language>: <Target Translated Text> \\ 

The evaluation score out of 100 is 
\end{mdframed}

\subsection{Example-based (In-Context Learning - ICL)}
In a similar manner, we identified and adjusted the prompt for example-based in-context learning with LLM. This specific prompt is outlined in the Example above (ICL Translation Prompt). In all of our experiments, specific to language pairs, we used a single and identical human translation evaluation with a score as a contextual learning example before executing the actual translation evaluation command using the training data mentioned in sub-section \ref{traindevingdata}.\\

\begin{mdframed}
\textbf{Example-based ICL Translation Evaluation Prompt Example:}\\

If the <Source Language> to <Target Language> translation score by human for '<Source Language Text>' to '<Target Language Translation>' is <Human Score> from 100 then following that, if you are an experienced translation evaluator and you need to evaluate a translation for <Source Language> language to <Target Language> language. \\ 

<Source Language>: <Source Language Text> \\ 

<Target Language>: <Target Translated Text> \\ 

The evaluation score out of 100 is 
\end{mdframed}

\section{Fine-tuning LLM for Reference-less Translation Evaluation}
To assess the potential enhancement in reference-less translation evaluation performance of LLMs beyond the zero-shot LLM baseline, we performed fine-tuning using the training corpus and method described below.

\begin{table}
\centering
\begin{tabular}{lll}
\toprule
\textbf{Method}                          & \textbf{Hyper-para}    & \textbf{Value}  \\
\midrule
\multirow{6}{*}{LoRA/Full}           & LoRA      &     PEFT\footnote{https://github.com/huggingface/peft}; FSDP\footnote{for Full Fine-Tuning}   \\
                                & rank              &    8    \\
                                & dropout           &    0.05    \\
                                & learning rate     &    1e-4    \\
                                & batch size &      4  \\
                                & epochs            &      5  \\
\bottomrule
\end{tabular}
\caption{Hyper-parameter configurations of LoRA based and full fine-tuning for 4*A100 40GB GPUs}
\label{LLM:Hyper-parameter}
\end{table}

\subsection{Human Judgement Train and Development Corpora}
\label{traindevingdata}

For this task, we take the publicly available datasets released under the quality estimation shared task \footnote{\url{https://wmt-qe-task.github.io/}} (MTQE) in WMT for Indian languages. English to five Indian languages are included in the shared task this year. The dataset comprises of 2 predominant language families: 3 from Indo-Aryan family and 2 Dravidian languages. The details of the train and development sets are presented in the table \ref{tab:data_det}.
\begin{table}[h]
    \centering
    \begin{tabular}{l|r|r}\hline
       \textbf{Lang Pair}  & \textbf{\#Train\_Snts} & \textbf{\#Dev\_Snts} \\\hline
        English-Hindi & 7000 & 1000 \\
        English-Gujarati & 7000 & 1000 \\
        English-Marathi & 26000 & 1000 \\
        English-Tamil & 7000 & 1000 \\
        English-Telugu & 7000 & 1000 \\\hline
    \end{tabular}
    \caption{Train and Development Set Details for Reference-less Translation Evaluation}
    \label{tab:data_det}
\end{table}
\subsubsection{Using Min-Max scaling on z-scores}
In order to fine-tune large language models on the aforementioned training data, we utilized the mean of z-scores\footnote{\url{https://www.investopedia.com/terms/z/zscore.asp}}, followed by language-specific min-max-based re-scaling \footnote{\url{https://en.wikipedia.org/wiki/Feature_scaling}} from 1 to 100 for the entire corpora.
% \usepackage{graphicx}
% \usepackage{lscape}
\begin{table*}[h]
\centering
\resizebox{2.05\columnwidth}{!}{%
\begin{tabular}{|l|ccc|ccc|ccc|ccc|ccc|}
\hline
\textbf{Model/Language Pair EN-} & \multicolumn{3}{c|}{\textbf{Hindi}} & \multicolumn{3}{c|}{\textbf{Gujarati}} & \multicolumn{3}{c|}{\textbf{Marathi}} & \multicolumn{3}{c|}{\textbf{Tamil}} & \multicolumn{3}{c|}{\textbf{Telugu}} \\\hline
\textbf{Co-relation scores} & \multicolumn{1}{l|}{\textbf{S}} & \multicolumn{1}{l|}{\textbf{P}} & \multicolumn{1}{l|}{\textbf{K}} & \multicolumn{1}{l|}{\textbf{S}} & \multicolumn{1}{l|}{\textbf{P}} & \multicolumn{1}{l|}{\textbf{K}} & \multicolumn{1}{l|}{\textbf{S}} & \multicolumn{1}{l|}{\textbf{P}} & \multicolumn{1}{l|}{\textbf{K}} & \multicolumn{1}{l|}{\textbf{S}} & \multicolumn{1}{l|}{\textbf{P}} & \multicolumn{1}{l|}{\textbf{K}} & \multicolumn{1}{l|}{\textbf{S}} & \multicolumn{1}{l|}{\textbf{P}} & \multicolumn{1}{l|}{\textbf{K}} \\ \hline \toprule 
\textbf{Llama-2-7b-Adpt} & \multicolumn{1}{c|}{\textbf{0.4268}} & \multicolumn{1}{c|}{\textbf{0.5303}} & 0.3056 & \multicolumn{1}{c|}{0.4795} & \multicolumn{1}{c|}{0.585} & 0.3757 & \multicolumn{1}{c|}{0.4481} & \multicolumn{1}{c|}{0.4939} & 0.3182 & \multicolumn{1}{c|}{0.5345} & \multicolumn{1}{c|}{\textbf{0.6632}} & 0.3967 & \multicolumn{1}{c|}{\textbf{0.3772}} & \multicolumn{1}{c|}{0.3324} & 0.2893 \\ \hline
\textbf{Llama-2-13b-Adpt} & \multicolumn{1}{c|}{0.3967} & \multicolumn{1}{c|}{0.587} & 0.2824 & \multicolumn{1}{c|}{\textbf{0.5226}} & \multicolumn{1}{c|}{\textbf{0.604}} & \textbf{0.419} & \multicolumn{1}{c|}{\textbf{0.5062}} & \multicolumn{1}{c|}{\textbf{0.5485}} & \textbf{0.3689} & \multicolumn{1}{c|}{0.5057} & \multicolumn{1}{c|}{0.6501} & 0.3774 & \multicolumn{1}{c|}{0.3408} & \multicolumn{1}{c|}{0.2976} & 0.2708 \\ \hline
\textbf{Mistral-7b-Adpt} & \multicolumn{1}{c|}{0.3923} & \multicolumn{1}{c|}{0.5295} & 0.2822 & \multicolumn{1}{c|}{0.4908} & \multicolumn{1}{c|}{0.5724} & 0.3994 & \multicolumn{1}{c|}{0.4736} & \multicolumn{1}{c|}{0.5518} & 0.3364 & \multicolumn{1}{c|}{0.5372} & \multicolumn{1}{c|}{0.615} & 0.3969 & \multicolumn{1}{c|}{0.333} & \multicolumn{1}{c|}{0.2685} & 0.2662 \\ \hline

\textbf{Llama-2-7b-Adpt-Full} & \multicolumn{1}{c|}{0.4003} & \multicolumn{1}{c|}{0.5249} & \textbf{0.3068} & \multicolumn{1}{c|}{0.4959} & \multicolumn{1}{c|}{0.5503} & 0.3958 & \multicolumn{1}{c|}{0.4524} & \multicolumn{1}{c|}{0.4938} & 0.3402 & \multicolumn{1}{c|}{\textbf{0.5413}} & \multicolumn{1}{c|}{0.6535} & \textbf{0.4138} & \multicolumn{1}{c|}{0.2940} & \multicolumn{1}{c|}{0.2500} & 0.2360 \\ \hline
\textbf{Llama-2-13b-Adpt-Full} & \multicolumn{1}{c|}{0.3004} & \multicolumn{1}{c|}{0.2164} & 0.2144 & \multicolumn{1}{c|}{0.3619} & \multicolumn{1}{c|}{0.3294} & 0.2994 & \multicolumn{1}{c|}{0.4424} & \multicolumn{1}{c|}{0.4229} & 0.3245 & \multicolumn{1}{c|}{0.4408} & \multicolumn{1}{c|}{0.484} & 0.3454 & \multicolumn{1}{c|}{0.2849} & \multicolumn{1}{c|}{0.1690} & 0.2271 \\ \hline

\textbf{Llama-2-7b-Adpt-Trans} & \multicolumn{1}{c|}{0.3273} & \multicolumn{1}{c|}{0.4602} & 0.2316 & \multicolumn{1}{c|}{0.4717} & \multicolumn{1}{c|}{0.5389} & 0.3824 & \multicolumn{1}{c|}{0.4196} & \multicolumn{1}{c|}{0.4554} & 0.2993 & \multicolumn{1}{c|}{0.4921} & \multicolumn{1}{c|}{0.6404} & 0.359 & \multicolumn{1}{c|}{0.2800} & \multicolumn{1}{c|}{0.2266} & 0.2137 \\ \hline
\textbf{Llama-2-13b-Adpt-Trans} & \multicolumn{1}{c|}{0.4011} & \multicolumn{1}{c|}{0.5165} & 0.2888 & \multicolumn{1}{c|}{0.5185} & \multicolumn{1}{c|}{0.6039} & 0.4088 & \multicolumn{1}{c|}{0.4451} & \multicolumn{1}{c|}{0.4956} & 0.3146 & \multicolumn{1}{c|}{0.5234} & \multicolumn{1}{c|}{0.6287} & 0.3892 & \multicolumn{1}{c|}{0.3264} & \multicolumn{1}{c|}{0.2714} & 0.2544 \\ \hline
\textbf{Mistral-7b-Adpt-Trans} & \multicolumn{1}{c|}{0.3507} & \multicolumn{1}{c|}{0.4611} & 0.2525 & \multicolumn{1}{c|}{0.4638} & \multicolumn{1}{c|}{0.5105} & 0.3671 & \multicolumn{1}{c|}{0.4308} & \multicolumn{1}{c|}{0.4849} & 0.3174 & \multicolumn{1}{c|}{0.468} & \multicolumn{1}{c|}{0.5796} & 0.3425 & \multicolumn{1}{c|}{0.3748} & \multicolumn{1}{c|}{\textbf{0.3403}} & \textbf{0.3042} \\ \hline 
\end{tabular}%
}
\caption{Correlation between Human judgement and Fine-tuned LLM with reference-less translation evaluation task for English to 4 Indian Languages is shown. Here, S, P, and K represent Spearman’s Rank Correlation Coefficient, Pearson Correlation Coefficient, and Kendall’s Rank Correlation Coefficient, respectively.}
\label{tab:FTresults}
\end{table*}
\subsection{Training Data for Translation Fine-Tuning}
\label{traintranslate}
To fine-tune LLMs for translation along with translation evaluation task under multi-task setting, we used the publicly available Bharat Parallel Corpus Collection (BPCC) designed for English to 22 Indic languages. We used BPCC-Human dataset, containing 2.2 million English-Indic pairs for this fine-tuning purpose.\\

\subsection{LLM Fine-tuning Details}
Considering the raw LLM performance, model parameters, and resource constraints, we selected a subset of LLMs for the fine-tuning process. Specifically, we opted for LLaMA-2-7b, LLaMA-2-13b, and Mistral-7B for the fine-tuning experiment. For these selected LLMs, we decided to perform low-rank adaptation-based fine-tuning \citep{hu2021lora} as well as full fine-tuning. In the fine-tuning process, we conducted a multi-lingual approach by considering corpora from all languages. Additionally, we conducted another experiment under multi-task settings, where we fine-tuned both LLaMA-2-7b and LLaMA-2-13b for reference-less translation evaluation and translation tasks. For the translation task, we utilized the translation dataset mentioned earlier (in subsection-\ref{traintranslate}), along with a simple JSON prompt instructing the model to translate a given sentence from the source language to the target language.\\

For both types of fine-tuning LLMs, we utilized the llama-recipes codebase\footnote{\url{https://github.com/facebookresearch/llama-recipes/}}  which provides an efficient implementation for LoRa-based adaptor fine-tuning with PEFT \citep{peft}. For more details, please refer to the llama-recipes documentation \footnote{\url{https://github.com/facebookresearch/llama-recipes/blob/main/docs/LLM_finetuning.md}}. The hyperparameters for the fine-tuning process are specified in Table \ref{LLM:Hyper-parameter}. 

\section{Results and Discussion}

In order to determine the performance of the LLM models, we examined the correlation between human judgments and metric output scores. This correlation served as the primary evaluation factor for identifying the most effective fine-tuned model. To conduct this analysis on the aforementioned development corpora, we utilized Spearman's Rank Correlation Coefficient, Pearson Correlation Coefficient, and Kendall's Rank Correlation Coefficient. To calculate these correlations, we employed the SciPy library\footnote{\url{https://scipy.org/}}.

\subsection{Zero shot vs ICL based Reference-less Translation Evaluation over Raw LLMs}
% \usepackage{graphicx}
\begin{table}[]
\centering
\resizebox{\columnwidth}{!}{%
\begin{tabular}{|r||r|r|r|r|r|}
\hline
\multicolumn{1}{|l||}{\textbf{Score Bin}} & \multicolumn{1}{l|}{\textbf{Hindi}} & \multicolumn{1}{l|}{\textbf{Gujarati}} & \multicolumn{1}{l|}{\textbf{Marathi}} & \multicolumn{1}{l|}{\textbf{Tamil}} & \multicolumn{1}{l|}{\textbf{Telugu}} \\ \hline\hline
99                                                                                                                       & 2                                           & 0                                              & 0                                             & 1                                           & 2                                            \\\hline
95                                                                                                                       & 5                                           & 43                                             & 32                                            & 9                                           & 0                                            \\\hline
90                                                                                                                       & 943                                         & 788                                            & 940                                           & 941                                         & 930                                          \\\hline
80                                                                                                                       & 50                                          & 155                                            & 27                                            & 45                                          & 68                                           \\\hline
70-80                                                                                                                    & 0                                           & 14                                             & 1                                             & 4                                           & 0                                            \\ \hline
\end{tabular}%
}
\caption{LLAMA-2-7B zero-shot performance on reference-less translation evaluation for English to 5 languages. Here, the `Score Bin' represents the score (out of 100) produced by LLAMA-2-7B. The corresponding language columns represent the respective sentence pairs that got that particular score bin. A total of 1000 samples were involved per language pair.}
\label{tab:zerobins}
\end{table}

Overall, the correlation scores with human evaluation for both Raw LLMs and In Context Learning (ICL) based LLMs in reference-less translation evaluation are substandard. All LLMs exhibited poor performance, which could be attributed to a lack of understanding or knowledge of the translation evaluation task. In our manual analysis, we observed that LLMs consistently provide a common number close to 100, regardless of the translation quality, for zero-shot evaluation (refer Table \ref{tab:zerobins}).  For in-context learning, LLMs simply mimic the example translation evaluation scores. Therefore, in response to our initial question, it can be concluded that \textbf{Raw LLMs do not possess inherent translation evaluation capabilities, either with zero-shot or example-driven contextual learning}. Refer to the appendix for examples and correlation details.

\subsection{Fine-Tuned LLM driven Reference-less Translation Evaluation}

We conducted an evaluation to compare the performance of our Fine-Tuned LLM models. The comparison results for English to 5 Indian language reference-less translation evaluation correlation with human judgments are presented in Table \ref{tab:FTresults}. It is evident that the highest correlations are achieved with LoRa-based fine-tuning methods for LLaMa-2-7b and LLaMa-2-13b models (indicated by -Adpt). On the other hand, Full Fine-tuning (indicated by -Adpt-Full) demonstrates maximum correlation scores for Tamil. However, overall, there is a low correlation for Telugu, indicating the need for further exploration, as Telugu language may have less representation compared to other languages in base LLM models. Additionally, it is worth noting that multi-task learning-based fine-tuning (indicated by -Adpt-Trans) does not lead to performance improvement compared to single-task fine-tuning for translation evaluation.\\

% \usepackage{graphicx}
\begin{table}[]
\centering
\resizebox{\columnwidth}{!}{%
\begin{tabular}{|l|rrr|}
\hline
\textbf{English to INs} & \multicolumn{3}{r|}{\textbf{Average Across Languages}} \\ \hline
\textbf{Co-relation scores} & \multicolumn{1}{r|}{\textbf{S}} & \multicolumn{1}{r|}{\textbf{P}} & \textbf{K} \\ \toprule \hline
\textbf{COMET-QE} & \multicolumn{1}{r|}{0.4568} & \multicolumn{1}{r|}{0.5236} & 0.32042 \\ \hline 
\textbf{Llama-2-7b-Adpt} & \multicolumn{1}{r|}{0.45322} & \multicolumn{1}{r|}{0.52096} & 0.3371 \\ \hline
\textbf{Llama-2-13b-Adpt} & \multicolumn{1}{r|}{\textbf{0.4574}} & \multicolumn{1}{r|}{\textbf{0.53744}} & \textbf{0.3437} \\ \hline
\textbf{Mistral-7b-Adpt} & \multicolumn{1}{r|}{0.44538} & \multicolumn{1}{r|}{0.50744} & 0.33622 \\ \hline
\textbf{Llama-2-7b-Adpt-Full} & \multicolumn{1}{r|}{0.43678} & \multicolumn{1}{r|}{0.4945} & 0.33852 \\ \hline
\textbf{Llama-2-13b-Adpt-Full} & \multicolumn{1}{r|}{0.36608} & \multicolumn{1}{r|}{0.32434} & 0.28216 \\ \hline
\textbf{Llama-2-7b-Adpt-Trans} & \multicolumn{1}{r|}{0.39814} & \multicolumn{1}{r|}{0.4643} & 0.2972 \\ \hline
\textbf{Llama-2-13b-Adpt-Trans} & \multicolumn{1}{r|}{0.4429} & \multicolumn{1}{r|}{0.50322} & 0.33116 \\ \hline
\textbf{Mistral-7b-Adpt-Trans} & \multicolumn{1}{r|}{0.41762} & \multicolumn{1}{r|}{0.47528} & 0.31674 \\ \hline

\end{tabular}%
}
\caption{The correlation between Human judgement and Fine-tuned LLM and COMET-QE with the reference-less translation evaluation task is shown, averaged across English to 4 Indian Languages (Hindi, Gujarati, Marathi, Tamil and Telugu). Here, S, P, and K represent Spearman’s Rank Correlation Coefficient, Pearson Correlation Coefficient, and Kendall’s Rank Correlation Coefficient, respectively.}
\label{tab:avgresults}
\end{table}

Table \ref{tab:avgresults} displays the average performance across all English to 4 language pairs for different fine-tuned LLMs and COMET-QE. We trained COMET-QE following the COMET architecture as described in COMET by Unbabel\footnote{\url{https://github.com/Unbabel/COMET}}, using the same training data configuration as our LLM fine-tuning. It is worth noting that the Llama-2-13b-Adpt model, which is an adapted model using the LORa method, achieves a high overall average human correlation across language pairs. This indicates the superior performance of LLM-driven reference-less translation evaluation. Hence, it highlights the potential of LLMs for the translation evaluation task involving English and Indian languages.\\

In answer to the question posed in the introduction, \textbf{Fine-tuning LLMs does indeed improve translation reference-less evaluation capabilities. However, when it comes to multi-task fine-tuning, including the translation task does not result in better performance compared to fine-tuning focused solely on the translation evaluation task.}\\

Based on our manual analysis of outputs, it became evident that the scores provided by the Llama-2-13b-Adpt model indicate superior quality. The model demonstrates the ability to detect translation differences, which is indeed reflected in its scores. Refer to the appendix for examples and correlation details.

\section{Limitations}
We conducted all our experiments using high-performance GPUs, specifically the A100-40GB, which may not be readily available for everyone to reproduce these experiments and obtain the same results due to compute limitations. To address this constraint, our goal is to make all outputs, including model outputs and results, openly accessible \footnote{\url{https://github.com/vmujadia/LLMT-Eval}} for further research.

\section{Conclusion and Future Work}
Our experiments and results indicate that fine-tuned LLMs show promise for translation evaluation in the targeted reference-less translation evaluation task. The findings call for further analysis and understanding of translation evaluation tasks involving large language models.\\

As part of future work, we plan to incorporate additional Indian languages for the reference-less translation evaluation task. Furthermore, we aim to utilize large language models for reference-driven translation evaluation, encompassing English and a wider range of Indian languages.\\

Overall, our study represents a significant milestone in assessing and enhancing the reference-less translation evaluation capabilities of LLMs, specifically in the context of English and Indian languages. For the future of automatic translation evaluation, we strongly advocate for solutions based on large language models as the primary method of evaluation. We have also released our models and predictions for future research.

\section*{Acknowledgement}

We would like to extend our sincere appreciation to Palash Gupta and Khoushik Ananth for their invaluable contributions at various stages of this project. The Ministry of Electronics and In- formation Technology, Government of India, has generously funded this endeavor, as part of the Sanction Order : 11(1)/2022-HCC(TDIL)-Part(2)/A/B/C and the Administrative Approval: 11(1)/2022-HCC(TDIL)-Part(2).

\bibliography{anthology,custom}
\onecolumn
\appendix
\section{Appendix}
\label{sec:appendix}
\subsection{Examples (English-Hindi)}
\begin{figure*}[!ht]
    \centering
    \includegraphics[width=.9\textwidth]{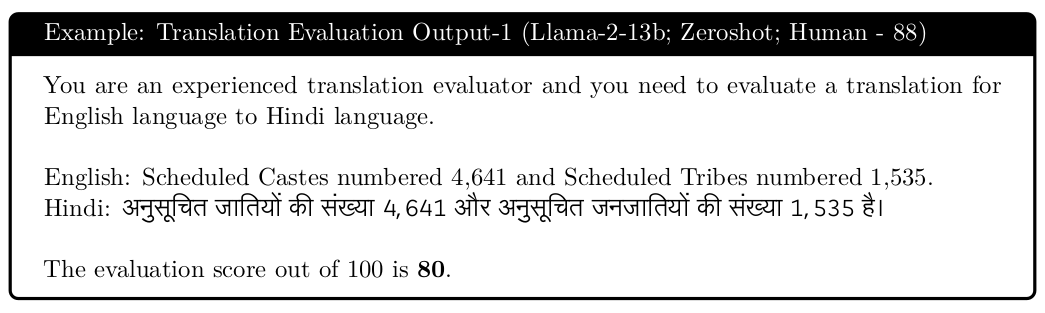}
    \label{fig:enter-label}
\end{figure*}

\begin{figure*}[!ht]
    \centering
    \includegraphics[width=.9\textwidth]{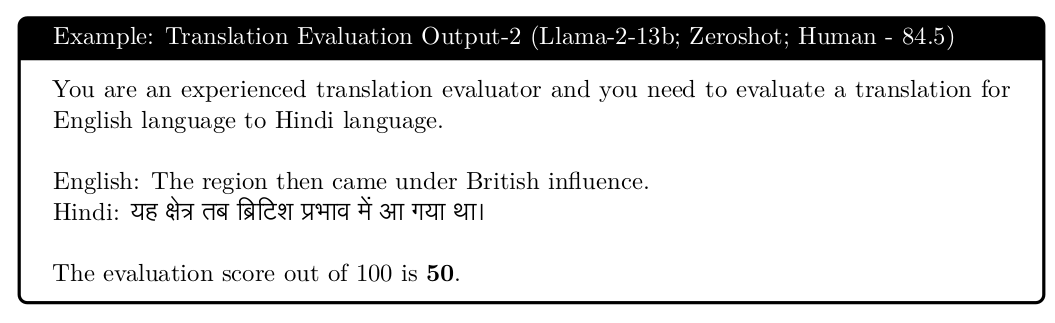}
    \label{fig:enter-label}
\end{figure*}

\begin{figure*}[!ht]
    \centering
    \includegraphics[width=.9\textwidth]{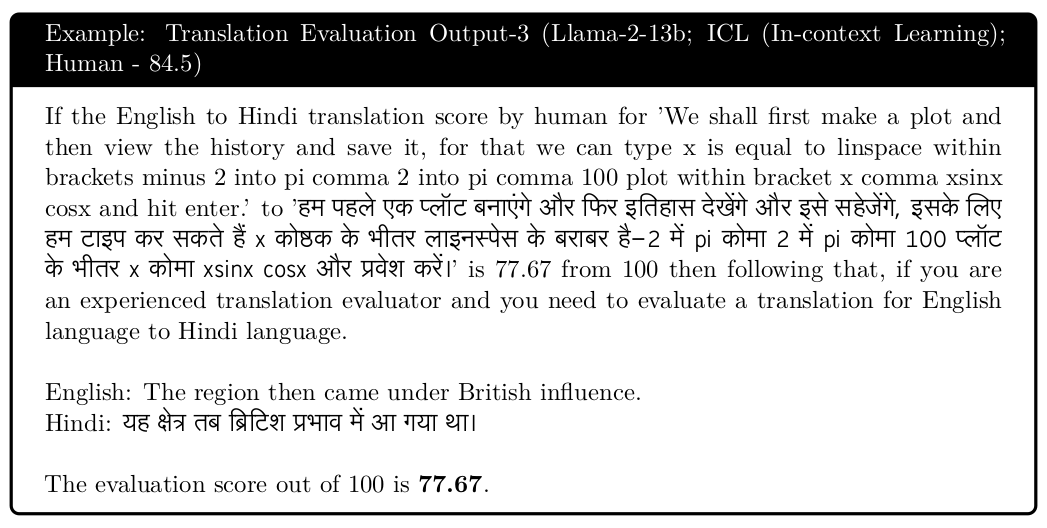}
    \label{fig:enter-label}
\end{figure*}

\begin{figure*}[!ht]
    \centering
    \includegraphics[width=.9\textwidth]{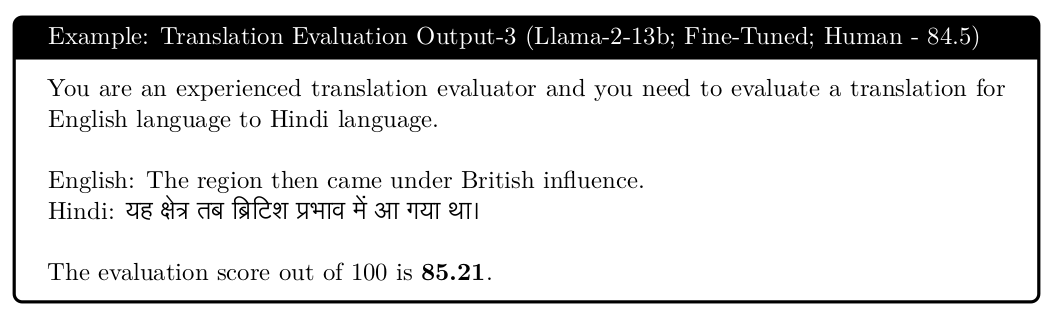}
    \label{fig:enter-label}
\end{figure*}

\begin{figure*}[!ht]
    \centering
    \includegraphics[width=.9\textwidth]{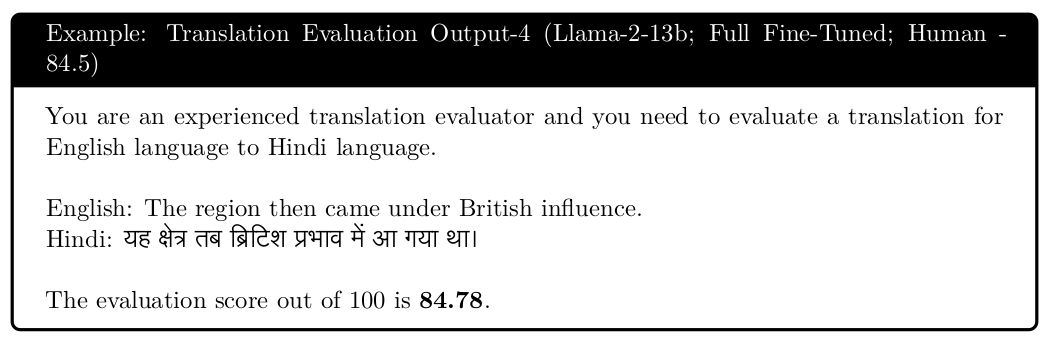}
    \label{fig:enter-label}
\end{figure*}

\begin{figure*}[!ht]
    \centering
    \includegraphics[width=.9\textwidth]{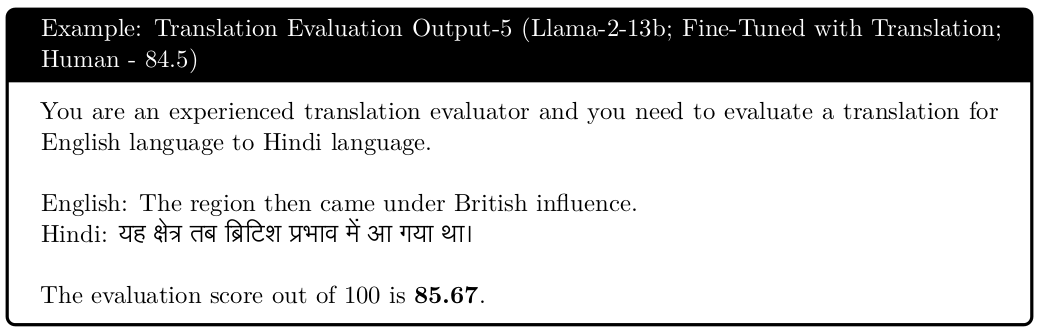}
    \label{fig:enter-label}
\end{figure*}

\begin{figure*}[!ht]
    \centering
    \includegraphics[width=.9\textwidth]{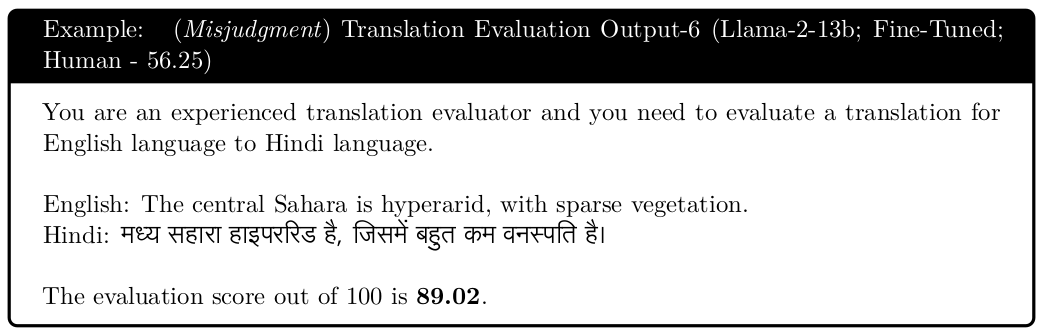}
    \label{fig:enter-label}
\end{figure*}
\subsection{Results}
\onecolumn
% \usepackage{graphicx}
% \usepackage{lscape}

\begin{table}[]
\centering
\resizebox{1.0\columnwidth}{!}{%
\begin{tabular}{|l|ccc|ccc|ccc|ccc|ccc|}
\hline
\textbf{Model/Language Pair: English-} & \multicolumn{3}{c|}{\textbf{Hindi}} & \multicolumn{3}{c|}{\textbf{Gujarati}} & \multicolumn{3}{c|}{\textbf{Marathi}} & \multicolumn{3}{c|}{\textbf{Tamil}} & \multicolumn{3}{c|}{\textbf{Telugu}} \\ \hline
\textbf{Co-relation scores} & \multicolumn{1}{l|}{\textbf{S}} & \multicolumn{1}{l|}{\textbf{P}} & \multicolumn{1}{l|}{\textbf{K}} & \multicolumn{1}{l|}{\textbf{S}} & \multicolumn{1}{l|}{\textbf{P}} & \multicolumn{1}{l|}{\textbf{K}} & \multicolumn{1}{l|}{\textbf{S}} & \multicolumn{1}{l|}{\textbf{P}} & \multicolumn{1}{l|}{\textbf{K}} & \multicolumn{1}{l|}{\textbf{S}} & \multicolumn{1}{l|}{\textbf{P}} & \multicolumn{1}{l|}{\textbf{K}} & \multicolumn{1}{l|}{\textbf{S}} & \multicolumn{1}{l|}{\textbf{P}} & \multicolumn{1}{l|}{\textbf{K}} \\ \hline
\textbf{bloom-7b1} & \multicolumn{1}{c|}{-0.0058} & \multicolumn{1}{c|}{-0.0116} & -0.0047 & \multicolumn{1}{c|}{0.0154} & \multicolumn{1}{c|}{0.0216} & 0.0129 & \multicolumn{1}{c|}{0.0122} & \multicolumn{1}{c|}{0.0222} & 0.0098 & \multicolumn{1}{c|}{-0.0256} & \multicolumn{1}{c|}{-0.015} & -0.0208 & \multicolumn{1}{c|}{0.0268} & \multicolumn{1}{c|}{0.0091} & 0.0222 \\ \hline
\textbf{falcon-7b} & \multicolumn{1}{c|}{0.0211} & \multicolumn{1}{c|}{0.0466} & 0.0162 & \multicolumn{1}{c|}{-0.0108} & \multicolumn{1}{c|}{0.0072} & -0.0079 & \multicolumn{1}{c|}{-0.1052} & \multicolumn{1}{c|}{-0.0083} & -0.0811 & \multicolumn{1}{c|}{0.0651} & \multicolumn{1}{c|}{0.0787} & 0.0496 & \multicolumn{1}{c|}{0.1151} & \multicolumn{1}{c|}{0.0775} & 0.0882 \\ \hline
\textbf{llama-7b} & \multicolumn{1}{r|}{0.02} & \multicolumn{1}{r|}{0.019} & \multicolumn{1}{r|}{0.0178} & \multicolumn{1}{r|}{-0.035} & \multicolumn{1}{r|}{-0.014} & \multicolumn{1}{r|}{-0.0276} & \multicolumn{1}{r|}{-0.0239} & \multicolumn{1}{r|}{-0.0243} & \multicolumn{1}{r|}{-0.019} & \multicolumn{1}{r|}{0.053} & \multicolumn{1}{r|}{0.0311} & \multicolumn{1}{r|}{0.0441} & \multicolumn{1}{r|}{0.0519} & \multicolumn{1}{r|}{0.0292} & \multicolumn{1}{r|}{0.0473} \\ \hline
\textbf{Llama-2-7b} & \multicolumn{1}{c|}{0.022} & \multicolumn{1}{c|}{0.0223} & 0.0177 & \multicolumn{1}{c|}{-0.0301} & \multicolumn{1}{c|}{-0.0135} & -0.0249 & \multicolumn{1}{c|}{-0.0222} & \multicolumn{1}{c|}{-0.0212} & -0.018 & \multicolumn{1}{c|}{0.0674} & \multicolumn{1}{c|}{0.0328} & 0.0547 & \multicolumn{1}{c|}{0.0606} & \multicolumn{1}{c|}{0.0304} & 0.0504 \\ \hline
\textbf{Mistral-7B-v0.1} & \multicolumn{1}{c|}{-0.0474} & \multicolumn{1}{c|}{-0.0582} & -0.0382 & \multicolumn{1}{c|}{0.0741} & \multicolumn{1}{c|}{0.063} & 0.0615 & \multicolumn{1}{c|}{0.077} & \multicolumn{1}{c|}{0.0766} & 0.0628 & \multicolumn{1}{c|}{-0.0487} & \multicolumn{1}{c|}{-0.054} & -0.0367 & \multicolumn{1}{c|}{-0.0143} & \multicolumn{1}{c|}{-0.0062} & -0.0116 \\ \hline
\textbf{opt-6.7b} & \multicolumn{1}{c|}{NAN} & \multicolumn{1}{c|}{NAN} & NAN & \multicolumn{1}{c|}{NAN} & \multicolumn{1}{c|}{NAN} & NAN & \multicolumn{1}{c|}{0.0826} & \multicolumn{1}{c|}{0.0229} & 0.0673 & \multicolumn{1}{c|}{NAN} & \multicolumn{1}{c|}{NAN} & NAN & \multicolumn{1}{c|}{NAN} & \multicolumn{1}{c|}{NAN} & NAN \\ \hline
\textbf{mpt-7b} & \multicolumn{1}{c|}{-0.0168} & \multicolumn{1}{c|}{-0.0003} & -0.0137 & \multicolumn{1}{c|}{0.0098} & \multicolumn{1}{c|}{0.0204} & 0.0082 & \multicolumn{1}{c|}{0.0234} & \multicolumn{1}{c|}{-0.0095} & 0.0191 & \multicolumn{1}{c|}{NAN} & \multicolumn{1}{c|}{NAN} & NAN & \multicolumn{1}{c|}{0.0405} & \multicolumn{1}{c|}{0.0312} & 0.0337 \\ \hline
\textbf{Llama-2-13b} & \multicolumn{1}{c|}{0.0347} & \multicolumn{1}{c|}{-0.0038} & 0.0267 & \multicolumn{1}{c|}{0.0862} & \multicolumn{1}{c|}{0.0965} & 0.0685 & \multicolumn{1}{c|}{-0.0691} & \multicolumn{1}{c|}{-0.0698} & -0.0551 & \multicolumn{1}{c|}{0.0748} & \multicolumn{1}{c|}{0.0276} & 0.0591 & \multicolumn{1}{c|}{0.0407} & \multicolumn{1}{c|}{-0.0077} & 0.0323 \\ \hline
\end{tabular}%
}
\caption{Correlation between Human judgement and Zero-shot LLM with reference-less translation evaluation task for English to 4 Indian Languages is shown. Here, S, P, and K represent Spearman’s Rank Correlation Coefficient, Pearson Correlation Coefficient, and Kendall’s Rank Correlation Coefficient, respectively.}
\label{tab:zeroresults}
\end{table}

% \usepackage{graphicx}
% \usepackage{lscape}
\begin{table}[]
\centering
\resizebox{1.0\columnwidth}{!}{%
\begin{tabular}{|l|ccc|ccc|ccc|ccc|ccc|}
\hline
\textbf{Model/Language Pair: English-} & \multicolumn{3}{c|}{\textbf{Hindi}} & \multicolumn{3}{c|}{\textbf{Gujarati}} & \multicolumn{3}{c|}{\textbf{Marathi}} & \multicolumn{3}{c|}{\textbf{Tamil}} & \multicolumn{3}{c|}{\textbf{Telugu}} \\ \hline
\textbf{Co-relation scores} & \multicolumn{1}{l|}{\textbf{S}} & \multicolumn{1}{l|}{\textbf{P}} & \multicolumn{1}{l|}{\textbf{K}} & \multicolumn{1}{l|}{\textbf{S}} & \multicolumn{1}{l|}{\textbf{P}} & \multicolumn{1}{l|}{\textbf{K}} & \multicolumn{1}{l|}{\textbf{S}} & \multicolumn{1}{l|}{\textbf{P}} & \multicolumn{1}{l|}{\textbf{K}} & \multicolumn{1}{l|}{\textbf{S}} & \multicolumn{1}{l|}{\textbf{P}} & \multicolumn{1}{l|}{\textbf{K}} & \multicolumn{1}{l|}{\textbf{S}} & \multicolumn{1}{l|}{\textbf{P}} & \multicolumn{1}{l|}{\textbf{K}} \\ \hline
\textbf{bloom-7b1} & \multicolumn{1}{c|}{-0.0496} & \multicolumn{1}{c|}{-0.0398} & -0.0405 & \multicolumn{1}{c|}{0.0093} & \multicolumn{1}{c|}{0.0075} & 0.0078 & \multicolumn{1}{c|}{0.1191} & \multicolumn{1}{c|}{0.1094} & 0.0973 & \multicolumn{1}{c|}{0.0233} & \multicolumn{1}{c|}{-0.0143} & 0.0191 & \multicolumn{1}{c|}{-0.0461} & \multicolumn{1}{c|}{-0.0381} & -0.0384 \\ \hline
\textbf{falcon-7b} & \multicolumn{1}{c|}{0.1374} & \multicolumn{1}{c|}{0.0175} & 0.1112 & \multicolumn{1}{c|}{0.03496} & \multicolumn{1}{c|}{0.0497} & 0.0284 & \multicolumn{1}{c|}{0.0008} & \multicolumn{1}{c|}{0.0009} & 0.0008 & \multicolumn{1}{c|}{0.127} & \multicolumn{1}{c|}{0.1015} & 0.102 & \multicolumn{1}{c|}{-0.1161} & \multicolumn{1}{c|}{-0.0825} & -0.0938 \\ \hline
\textbf{llama-7b} & \multicolumn{1}{c|}{-0.0492} & \multicolumn{1}{c|}{-0.0615} & -0.0402 & \multicolumn{1}{c|}{NAN} & \multicolumn{1}{c|}{NAN} & NAN & \multicolumn{1}{c|}{NAN} & \multicolumn{1}{c|}{NAN} & NAN & \multicolumn{1}{c|}{-0.0416} & \multicolumn{1}{c|}{-0.0262} & -0.034 & \multicolumn{1}{c|}{NAN} & \multicolumn{1}{c|}{NAN} & NAN \\ \hline
\textbf{Llama-2-7b} & \multicolumn{1}{c|}{NAN} & \multicolumn{1}{c|}{NAN} & NAN & \multicolumn{1}{c|}{NAN} & \multicolumn{1}{c|}{NAN} & NAN & \multicolumn{1}{c|}{NAN} & \multicolumn{1}{c|}{NAN} & NAN & \multicolumn{1}{c|}{-0.0416} & \multicolumn{1}{c|}{-0.0262} & -0.034 & \multicolumn{1}{c|}{NAN} & \multicolumn{1}{c|}{NAN} & NAN \\ \hline
\textbf{Mistral-7B-v0.1} & \multicolumn{1}{c|}{0.0382} & \multicolumn{1}{c|}{0.0144} & 0.0312 & \multicolumn{1}{c|}{0.0287} & \multicolumn{1}{c|}{0.0382} & 0.0239 & \multicolumn{1}{c|}{NAN} & \multicolumn{1}{c|}{NAN} & NAN & \multicolumn{1}{c|}{-0.0023} & \multicolumn{1}{c|}{-0.0117} & -0.0018 & \multicolumn{1}{c|}{0.008} & \multicolumn{1}{c|}{0.0348} & 0.0066 \\ \hline
\textbf{opt-6.7b} & \multicolumn{1}{c|}{-0.0576} & \multicolumn{1}{c|}{-0.0436} & -0.0471 & \multicolumn{1}{c|}{-0.0095} & \multicolumn{1}{c|}{0.0005} & -0.0079 & \multicolumn{1}{c|}{0.0278} & \multicolumn{1}{c|}{0.0129} & 0.0227 & \multicolumn{1}{c|}{0.1141} & \multicolumn{1}{c|}{0.0936} & 0.0933 & \multicolumn{1}{c|}{0.0918} & \multicolumn{1}{c|}{0.0791} & 0.0764 \\ \hline
\textbf{mpt-7b} & \multicolumn{1}{c|}{0.0661} & \multicolumn{1}{c|}{0.0454} & 0.054 & \multicolumn{1}{c|}{-0.0539} & \multicolumn{1}{c|}{-0.0498} & -0.0449 & \multicolumn{1}{c|}{0.0639} & \multicolumn{1}{c|}{0.0431} & 0.0522 & \multicolumn{1}{c|}{0.1546} & \multicolumn{1}{c|}{0.137} & 0.1263 & \multicolumn{1}{c|}{-0.1268} & \multicolumn{1}{c|}{-0.0826} & -0.1054 \\ \hline
\textbf{Llama-2-13b} & \multicolumn{1}{c|}{NAN} & \multicolumn{1}{c|}{NAN} & NAN & \multicolumn{1}{c|}{NAN} & \multicolumn{1}{c|}{NAN} & NAN & \multicolumn{1}{c|}{NAN} & \multicolumn{1}{c|}{NAN} & NAN & \multicolumn{1}{c|}{-0.0006} & \multicolumn{1}{c|}{-0.0167} & -0.0005 & \multicolumn{1}{c|}{0.0159} & \multicolumn{1}{c|}{0.0214} & 0.0132 \\ \hline
\end{tabular}%
}
\caption{Correlation between Human judgement and In Context Learning LLM with reference-less translation evaluation task for English to 4 Indian Languages is shown. Here, S, P, and K represent Spearman’s Rank Correlation Coefficient, Pearson Correlation Coefficient, and Kendall’s Rank Correlation Coefficient, respectively.}
\label{tab:ICLresults}
\end{table}
\end{document}